\documentclass[11pt,a4paper]{article}
\usepackage[hyperref]{acl2020}
\usepackage{times}
\usepackage{latexsym}

\usepackage{microtype}

\aclfinalcopy % Uncomment this line for the final submission
\def\aclpaperid{1205} %  Enter the acl Paper ID here

\usepackage{graphicx}
\usepackage{multirow}
\usepackage{times}
\usepackage{latexsym}
\usepackage{amsmath,amsthm}
\usepackage{amssymb}
\usepackage{amsfonts}
\usepackage{algorithm}
\usepackage{mathrsfs}
\usepackage[noend]{algpseudocode}
\usepackage{booktabs} % For formal tables
\usepackage{color}
\usepackage{natbib}
\usepackage{hyperref} %It's Incompatible for AAAI 
\usepackage{url}
\usepackage{caption} %It's Incompatible for AAAI 
\usepackage{algpseudocode}
\usepackage{pifont}
\usepackage{bm}
\usepackage{comment}

\def\1{{\bf{1}}}
\def\0{{\bf{0}}}

\def\c{{\bf c}}

\def\e{{\bf e}}

\def\r{{\bf r}}

\def\t{{\bf t}}

\def\E{{\bf E}}

\def\J{{\bf J}}

\def\M{{\bf M}}

\def\R{{\bf R}}
\def\S{{\bf S}}
\def\T{{\bf T}}
\def\U{{\bf U}} 
\def\V{{\bf V}}

\def\Dcal{{\mathcal{D}}}
\def\Ecal{{\mathcal{E}}}

\def\Hcal{{\mathcal{H}}}

\def\Rcal{{\mathcal{R}}}

\def\Tcal{{\mathcal{T}}}

\def\Zcal{{\mathcal{Z}}}

\def\Rbb{{\mathbb R}}

\newcommand{\hytt}[1]{\texttt{\hyphenchar\font=\defaulthyphenchar #1}}

\def\cites #1{{\citeauthor{#1} \shortcite{#1}}}  %this command is defined to cite reference as subject in text.
\newtheorem{theorem}{Mechanism}

\title{Connecting Embeddings for Knowledge Graph Entity Typing}

\author{Yu Zhao\textsuperscript{1,}\thanks{\ \ Equal Contribution. Corresponding author: Y. Zhao (zhaoyu@swufe.edu.cn).}, Anxiang Zhang\textsuperscript{2,*}, Ruobing Xie\textsuperscript{3}, Kang Liu\textsuperscript{4,5}, Xiaojie Wang\textsuperscript{6}\\
  \textsuperscript{1}Fintech Innovation Center, School of Economic Information Engineering, \\ Southwestern University of Finance and Economics, Chengdu, China \\
  \textsuperscript{2}School of Computer Science, Carnegie Mellon University, Pittsburgh, USA \\
 \textsuperscript{3}WeChat Search Application Department, Tencent, Beijing, China \\
 \textsuperscript{4}National Laboratory of Pattern Recognition (NLPR), Institute of  Automation, \\
 Chinese Academy of Sciences, Beijing, 100190, China \\
  \textsuperscript{5}University of Chinese Academy of Sciences, Beijing, 100049, China\\
 \textsuperscript{6} School of Computer Science, Beijing University of Posts and \\ Telecommunications, Beijing, China}

\date{}

\begin{document}
\maketitle
\begin{abstract}
Knowledge graph (KG) entity typing aims at inferring possible missing entity type instances in KG, which is a very significant but still under-explored subtask of knowledge graph completion. In this paper, we propose a novel approach for KG entity typing which is trained by jointly utilizing {\emph{local typing knowledge}} from existing entity type assertions and {\emph{global triple knowledge}} from KGs. Specifically, we present two distinct knowledge-driven effective mechanisms of entity type inference. Accordingly, we build two novel embedding models to realize the mechanisms. Afterward, a joint model with them is used to infer missing entity type instances, which favors inferences that agree with both entity type instances and triple knowledge in KGs. Experimental results on two real-world datasets (Freebase and YAGO) demonstrate the effectiveness of our proposed mechanisms and models for improving KG entity typing. The source code and data of this paper can be obtained from: \url{https://github.com/Adam1679/ConnectE}
\end{abstract}

\section{Introduction}
The past decade has witnessed great thrive in building web-scale knowledge graphs (KGs), such as Freebase \cite{Bollacker:Freebase}, YAGO \cite{Suchanek:Yago}, Google Knowledge Graph \cite{Dong:Knowledge}, which usually consists of a huge amount of triples in the form of ({\emph{head entity}}, {\emph{relation}}, {\emph{tail entity}}) (denoted ($ e,r,\tilde{e}$)). KGs usually suffer from \emph{incompleteness} and miss important facts, jeopardizing their usefulness in downstream tasks such as question answering \cite{Elsahar:Zero2018}, semantic parsing \cite{Berant:Semantic}, relation classification \cite{Zeng2014Relation}. Hence, the task of knowledge graph completion (KGC, i.e. completing knowledge graph entries) is extremely significant and attracts wide attention. 

This paper concentrates on KG entity typing, i.e. inferring missing entity type instances in KGs, which is an important sub-problem of KGC. Entity type instances, each of which is in the formed of (\emph{entity, entity type}) (denoted ($e,t$)), are essential entries of KGs and widely used in many NLP tasks such as relation extraction \cite{Zhang:Embedding2018,Jain:Type2018}, coreference resolution \cite{Hajishirzi:Joint}, entity linking \cite{Gupta:Entity2017}. Most previous works of KGC focus on inferring missing entities and relationships \cite{Bordes:Translating,Wang:Knowledge2014,Lin:Learning,Dettmers:Convolutional,Ding:Improving,Nathani2019Learning}, paying less attention to entity type prediction. 
However, KGs also usually suffer from entity types incompleteness. For instance, 10\% of entities in \emph{FB15k} \cite{Bordes:Translating}, which have the \emph{/music/artist} type, miss the \emph{/people/person} type \cite{Moon:Learning}. KG entity type incompleteness leads to some type-involved algorithms in KG-driven tasks grossly inefficient or even unavailable.

\begin{figure}[t]
    \centering
    \includegraphics[width=0.45\textwidth]{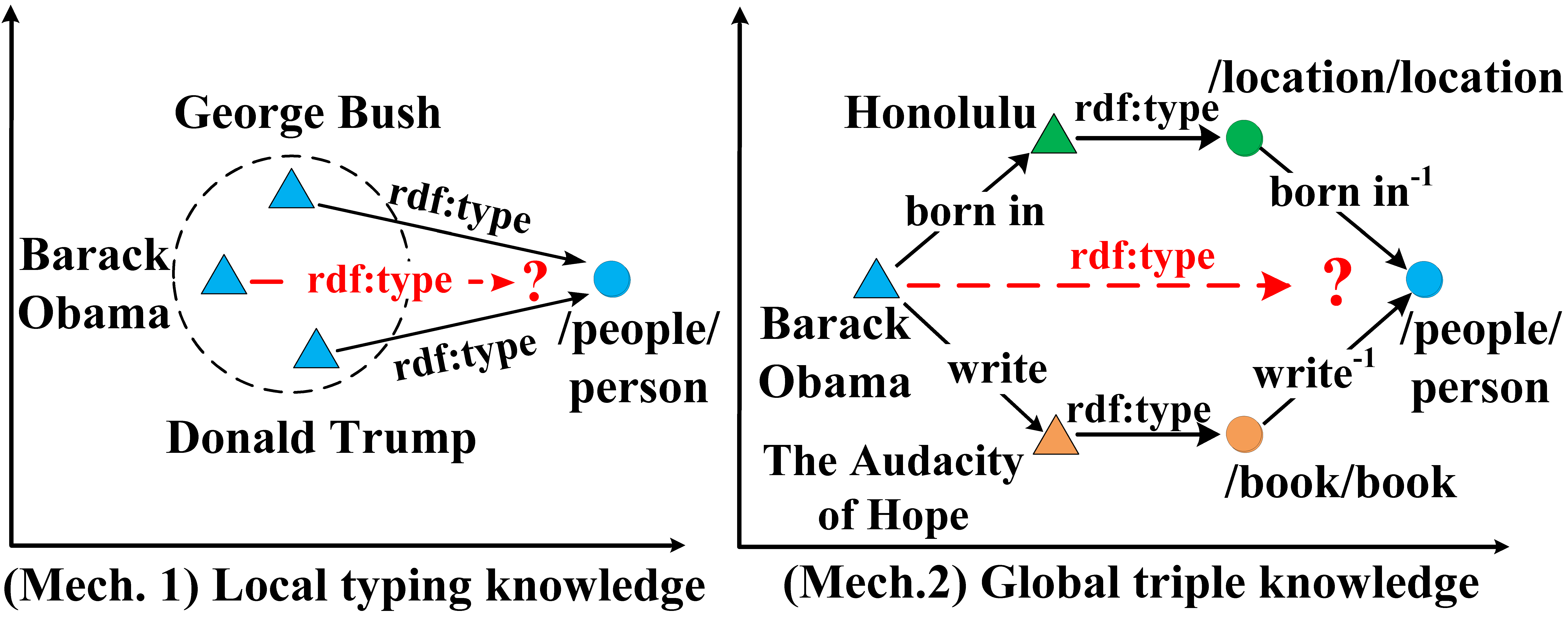}
    \caption{Effective mechanisms of entity type inference with local typing knowledge and global triple knowledge.}
    \label{fig:motivation}
\end{figure}

To solve {KG entity type incompleteness} issue, in this paper we propose a novel embedding methodology to infer missing entity type instances that employs not only \emph{local typing knowledge} from entity type assertions, as most conventional models do, but also leverages \emph{global triple knowledge} from KGs. Accordingly, we build two distinct knowledge-driven type inference mechanisms with these two kinds of structural knowledge.
\begin{theorem}
\label{machanism-E2T}
\textbf{Missing entity types of an entity can be found from other entities that are close to the entity in the embedding space, using local typing knowledge as in Fig.$\ref{fig:motivation}$(Mech.1).}
\end{theorem}
\begin{theorem}
\label{machanism-TRT}
\textbf{Missing entity types of an (head or tail) entity can be inferred from the types of other (tail or head) entities through their relationships, using global triple knowledge as in Fig.$\ref{fig:motivation}$(Mech.2).}
\end{theorem}
\noindent{The} main idea behind Mech.$\ref{machanism-E2T}$ is based on the observation that the learned entities' embeddings by conventional KG embedding methods \cite{Ji2016Knowledge,Xie:Representation_type} cluster well according to their types in vector space. For instance, in Fig.$\ref{fig:motivation}$(Mech.1), given an entity \emph{Barack Obama}, it's missing hierarchical type \emph{/people/person} can be induced by the given hierarchical type of similar entity \emph{Donald Trump}. In addition, the key motivation behind Mech.$\ref{machanism-TRT}$ is that the relationship shall remain unchanged if the entities in a triple fact are replaced with their corresponding hierarchical types. For instance, given a global triple fact (\textit{Barack Obama, \underline{born$\_$in}, Honolulu}), under this assumption, we can induce a new type triple (\textit{/people/person, \underline{born$\_$in}, /location/location})\footnote{For more clarity, we represent it as (\textit{/location/location, born\_in$^{-1}$, /people/person}) in Fig.$\ref{fig:motivation}$(Mech.2).}. Formally, $\vec{\textbf{\emph{Honolulu}}} - \vec{\textbf{\emph{Barack Obama}}} = \vec{\textbf{\emph{/location/location}}} - \vec{\textbf{\emph{/people/person}}}$ (= $\vec{\textbf{\emph{born\_in}}}$), which can be used to infer missing entity types, e.g. (\emph{Barack Obama, type=? }) via $\vec{{\emph{Barack Obama}}} - \vec{{\emph{Honolulu}}} + \vec{{\emph{/location/location}}}$ = $\vec{{\emph{/people/person}}}$, as Mech.$\ref{machanism-TRT}$ does. Fig.$\ref{fig:motivation}$ demonstrates a simple illustration of effective mechanisms of entity type inference. Both mechanisms are utilized to build our final composite model. 

Specifically, we build two embedding models to realize the two mechanisms respectively. First, considering entities and entity types are completely distinct objects, we build two distinct embedding spaces for them, i.e., {\textbf{entity space}} and {\textbf{entity type space}}. Accordingly, we encode $(e,t)$ entity type instance by projecting the entity from entity space to entity type space with mapping matrix $\M$, hence we have (1): \fbox{$\M\cdot \e \simeq \t$}, called {\bf E2T}.
Moreover, we learn the plausibility of ($t_e, r, t_{\tilde{e}}$) global type triple by newly generalizing from ($e,r,\tilde{e}$) global triple fact, even though this type triple is not present originally.
Following translating assumption \cite{Bordes:Translating}, we have (2): \fbox{$\t_{\tilde{e}} - \r^\circ \simeq \t_e$}, called {\bf TRT}. 
E2T and TRT are the implementation models of the two mechanisms. Fig.$\ref{fig:E2T-TRT}$ demonstrates a brief illustration of our models. A ranking-based embedding framework is used to train our models. 
Thereby, entities, entity hierarchical types, and relationships are all embedded into low-dimensional vector spaces, where the composite energy score of both E2T and TRT are computed and utilized to determine the optimal types for (\emph{entity, entity type}=?) incomplete assertions. The experimental results on real-world datasets show that our composite model achieves significant and consistent improvement compared to all baselines in entity type prediction and achieves comparable performance in entity type classification. 

{\bf Our contributions} are as follows:
\begin{itemize}
    \item We propose a novel framework for inferring missing entity type instances in KGs by connecting entity type instances and global triple information and correspondingly present two effective mechanisms.
    \item Under these mechanisms, we propose two novel embedding-based models: one for predicting entity types given entities and another one to encode the interactions among entity types and relationships from KGs. A combination of both models are utilized to conduct entity type inference.
    \item We conduct empirical experiments on two real-world datasets for entity type inference, which demonstrate our model can successfully take into account global triple information to improve KG entity typing.
\end{itemize}
\begin{table*}[htb]\scriptsize
\caption{Entity type embedding models. }
\label{related-models}
\newcommand{\tabincell}[2]{\begin{tabular}{@{}#1@{}}#2\end{tabular}}
\centering
\begin{tabular}[t]{l||c|c|c|c|c}
\toprule
\multirow{2}*{\bf{Models} } &\multicolumn{2}{c|}{\bf Energy function} &\multirow{2}*{\bf Parameters}&\multirow{2}*{\bf Sources}& \multirow{2}*{\tabincell{c}{ \bf Training \\ \bf strategy}}\\
\cline{2-3}
       &\boldsymbol{$\S_{e2t}(e,t)$ }&\boldsymbol{$\S_{triple}(\cdot)$ }&&& \\
\midrule
LM  \cite{Neelakantan:Inferring}&$\e^\top \t$&  N/A &$\e,\t \in \Rbb^{\kappa}$&entity type instances& N/A \\
\midrule
PEM  \cite{Neelakantan:Inferring}&$\e^\top \U \V^\top \t$ & N/A &\tabincell{c}{ $\e \in \Rbb^{\kappa}$, $\t \in \Rbb^{\ell}$,\\ $\U \in \Rbb^{\kappa \times d}$,$\V \in \Rbb^{\ell \times d}$} &entity type instance&N/A\\
\midrule
RESCAL \cite{Nickel:A2011} &N/A&$\e^\top\M_r\boldsymbol{\tilde{e}}$&\tabincell{c}{ $\e,\boldsymbol{\tilde{e}} \in \Rbb^{\kappa}$, $\M_r\in \Rbb^{\kappa \times \kappa}$ }&mixed triple knowledge & syn.\\
\midrule
RESCAL-ET \cite{Moon:Learning} &$\|\e - \t\|_1$&$\e^\top\M_r\boldsymbol{\tilde{e}}$&\tabincell{c}{ $\e,\boldsymbol{\tilde{e}}, \t \in \Rbb^{\kappa}$, $\M_r\in \Rbb^{\kappa \times \kappa}$ }&entity type inst./ triple know. & asyn.\\
\midrule
HOLE \cite{Nickel:Holographic} &N/A&$\r^\top(\e\star\boldsymbol{\tilde{e}})$& $\e,\r,\boldsymbol{\tilde{e}} \in \Rbb^{\kappa}$ &mixed triple knowledge & syn.\\
\midrule
HOLE-ET \cite{Moon:Learning} &$\|\e - \t\|_1$&$\r^\top(\e\star\boldsymbol{\tilde{e}})$& $\e,\r,\boldsymbol{\tilde{e}}, \t \in \Rbb^{\kappa}$ &entity type inst./ triple know. & asyn.\\
\midrule
TransE \cite{Bordes:Translating}&N/A&$\|\e +\r -\boldsymbol{\tilde{e}}\|$& $\e,\r,\boldsymbol{\tilde{e}} \in \Rbb^{\kappa}$ &mixed triple knowledge & syn.\\
\midrule
TransE-ET \cite{Moon:Learning}&$\|\e - \t\|_1$&$\|\e +\r -\boldsymbol{\tilde{e}}\|$& $\e,\r,\boldsymbol{\tilde{e}}, \t \in \Rbb^{\kappa}$ &entity type inst./ triple know. & asyn.\\
\midrule
ETE \cite{Moon:Learning} &$\|\e - \t\|_1$&$\|\e +\boldsymbol{\tilde{e}} + \c - \r\|$ &$\e,\r,\boldsymbol{\tilde{e}} ,\c, \t \in \Rbb^{\kappa}$ &entity type inst./ triple know. & asyn.\\
\midrule
\tabincell{c}{\bf ConnectE \bf (our proposed)} &$\| \M \cdot \e -  \t \|_2^2$&\tabincell{c}{$\| \e + \r^\star - \boldsymbol{\tilde{e}} \|_2^2$ , \\$\| \t_e + \r^\circ - \t_{\tilde{e}} \|_2^2$ }&\tabincell{c}{ $\e,\r^\star \in \Rbb^\kappa, \t, \r^\circ \in \Rbb^\ell$,\\ $\M \in \Rbb^{\ell \times \kappa}$ }&entity type inst./ triple know. & syn.\\
\bottomrule
\end{tabular}
\end{table*}

\section{Related Works}
\label{relatedworks}
Entity typing is valuable for many NLP tasks \cite{Yaghoobzadeh2018Corpus-level}, such as knowledge base population \cite{Zhou:Zero:2018}, question answering \cite{Elsahar:Zero2018}, etc. In recent years, researchers attempt to mine fine-grained entity types \cite{Yogatama:Embedding2015,Choi:Ultra2018,Xu:Neural2018,Yuan:Otyper:2018} with external text information, such as web search query logs \cite{Pantel:Mining}, the textual surface patterns \cite{Yao:Universal}, context representation \cite{Abhishek:Fine:2017}, Wikipedia \cite{Zhou:Zero:2018}. Despite their success, existing methods rely on additional external sources, which might not be feasible for some KGs.

To be more universal, \cites{Neelakantan:Inferring} propose two embedding models, i.e. linear model (LM) and projection embedding model (PEM), which can infer missing entity types only with KG itself. Although PEM has more expressive power than LM, however, both of them ignore global triple knowledge, which could also be helpful for encoding entity type assertions via shared entities' embeddings. To address this issue, \cites{Moon:Learning} propose a state-of-the-art model (ETE) to combine triple knowledge and entity type instances for entity type prediction, and build two entity type embedding methodologies: (1) Synchronous training: treat (\emph{entity, entity type}) assertions as special triple facts that have a unique relationship ``\emph{rdf:type}", e.g. (\textit{Barack Obama, ``rdf:type", person}), and encode all mixed triple facts (original triple data fused with all generated special ones) by conventional entity relation embedding models, such as RESCAL \cite{Nickel:A2011}, HOLE \cite{Nickel:Holographic} and TransE \cite{Bordes:Translating}. (2) Asynchronous training: first learn the entities' embeddings $\e$ by conventional entity relation embedding models mentioned above, and then only update entity types' embeddings $\t$ for $\min \|\e-\t\|_{\ell1}$ while keeping $\e$ fixed, called RESCAL-ET, HOLE-ET, TransE-ET and ETE.  Although these approaches expect to explore global triple knowledge for entity type prediction, they still lack of expressive ability due to its simplicity of embeddings. In addition, they irrationally assume both the embeddings of entities and entity types being in the same latent space ($\in \mathbb{R}^\kappa$). Since entities and entity types are completely distinct objects, it may not be reasonable to represent them in a common semantic space. 

In this paper, we introduce an enhanced KG entity type embedding model with better expressing and reasoning capability considering both local entity typing information and global triple knowledge in KGs. Note that incorporating more external information \cite{Jin2018Attributed,Neelakantan:Inferring} is not the main focus in this paper, as we only consider the internal structural information in KGs instead, which correspondingly makes our work much more challenging but also more universal and flexible due to the limited information. Recently, \cite{Lv2018Differentiating,Hao2019Universal} also attempt to embedding structural information in KG. However, the goals and models are very different from ours. They encodes the “concepts”, not hierarchical types. On the contrary, we focus on the latter not the former. Table $\ref{related-models}$ summarizes the energy functions and other different settings of entity type embedding models. 

\section{Embedding-based Framework}
\label{methodology}
We consider a KG containing entity type instances of the form $(e, t) \in \Hcal$ ($\Hcal$ is the training set consists of lots of (\emph{entity, entity type}) assertions), where $e \in \Ecal$ ($\Ecal$ is the set of all entities) is an entity in the KG with the type $t \in \Tcal$ ($\Tcal$ is the set of all types). For example, $e$ could be \emph{Barack Obama} and $t$ could be \emph{/people/person}. As a single entity can have multiple types, entities in KG often miss some of their types. The aim of this work is to infer missing entity type instances in KGs.

Our work concerns energy-based methods, which learn low-dimensional vector representations (\emph{embeddings}) of atomic symbols (i.e. \emph{entities, entity hierarchical types, relationships}). In this framework, we learn two submodels: (1) one for predicting entity types given entities, and (2) another one to encode the interactions among entity types and relationships from KGs. The joint action of both models in prediction allows us to use the connection between triple knowledge and entity type instances to perform KG entity typing. 

\begin{figure}[t]
    \centering
    \includegraphics[width=0.45\textwidth]{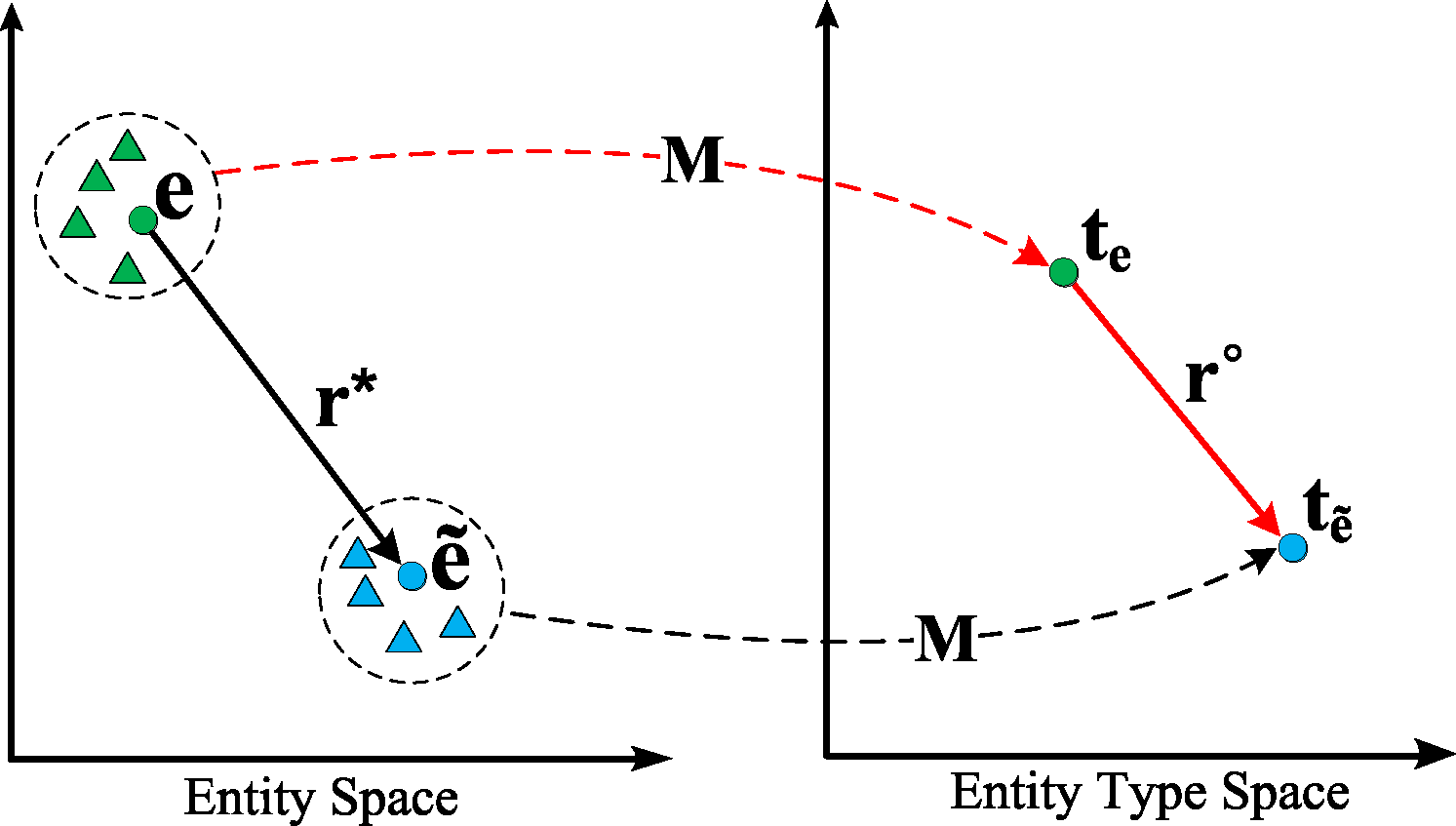}
    \caption{Simple illustration of E2T and TRT.}
    \label{fig:E2T-TRT}
\end{figure}

\subsection{E2T: Mapping Entities to Types}
The first model (E2T) of the framework concerns the learning of a function $\S_{e2t}(e, t)$ with local typing knowledge from entity type instances, which is designed to score the similarity of an entity $e$ and a type $t$. The main ideas behind this model are as follows:
(1) Since the learned entity embeddings cluster well when they have the same or similar types, therefore, it is rather intuitive that the entity type embedding represents the projective common concept representation of a cluster of entities, i.e., $f_{proj}(\e) \simeq \t_e, \forall e \in \Ecal$. $\e$ $( \in \mathbb{R}^\kappa)$ is the embedding of the entity $e$, $\t_e$ $(\in \Rbb^{\ell})$ is the embedding of the type $t_e$. The entity type embedding represents common information of their entities, it thus should have fewer variates, i.e., $\ell < \kappa$. 
(2) Since the entities and entity types are totally distinct objects, we respectively build two embedding space for them, i.e., {\bf entity space} and {\bf entity type space}. 
(3) Inspired by the previous work TranSparse \cite{Ji2016Knowledge} projecting entities from entity space to relation space with operation matrix $\M$, which we adapted, replacing relation space with entity type space, we thus define $f_{proj}(\e) = \M\cdot \e\ ( \simeq \t_e)$. Therefore, this model consists of first projecting entity embedding into entity type space, and then computing a similarity measure between this projection and an entity type embedding. The scoring function of E2T given ($e,t$) is:
\begin{equation}
    \S_{e2t}(e, t) = \| \M \cdot \e -  \t \|_{\ell2}^2 \ ,
\end{equation}
where $\M \in \mathbb{R}^{\ell \times \kappa}$ is a transfer matrix mapping entity embeddings into entity type space. The score is expected to be lower for a golden entity type instance and higher for an incorrect one. 

\subsection{TRT: Encoding Triples in KGs}
Using only entity type instances for training ignores much of relational knowledge that can leverage from triple facts in KGs. In order to connect this relational data with our model, we propose to learn entity type and relationship embeddings from global triple knowledge from KGs. The key motivations behind this model are: (1)  As mentioned above, the entities cluster well according to their types. Therefore, we believe that an essential premise of a triple (\emph{head entity, relationship, tail entity}) holds is that its corresponding entity types should first conform to this relationship. Hence, we can build a new {\bf entity type triple} (\emph{head type, relationship, tail type}) by replacing both head entity and tail entity with their corresponding types: i.e. ${ {(e, r, \tilde{e})}}$ $\stackrel{replace}{\longrightarrow}$ ${ {(t_e, r, t_{\tilde{e}})}}$. 
$(e, r, \tilde{e}) \in \Dcal $, $\Dcal$ is the training set consists of a lot of triples. $r \in \Rcal$ ($\Rcal$ is the set of relationships). $t_e$ and $t_{\tilde{e}}$ stand for the hierarchical types of left entity $e$ and right entity $\tilde{e}$ respectively. 
(2) Since the relationship $r$ remains unchanged in replacement, we build two differentiated embeddings for the $i$-th relationship $r_i$ in two embedding spaces:  $\r_i^\star (\in \Rbb^\kappa)$ in entity space and $\r_i^\circ$ $(\in \Rbb^\ell)$ in entity type space. 
(3) Given entity type triple ${ {(t_e, r, t_{\tilde{e}})}}$, under translation assumption \footnote{We chose TransE in this paper, and it is not difficult for other enhanced translation-based methods to model triple knowledge, such as Trans(H, R, D and G) \cite{Wang:Knowledge2017}.} as in \cite{Bordes:Translating}, we have: $\t_{\tilde{e}} - \r^\circ \simeq \t_e$. Hence, the scoring function is defined as:
\begin{equation}
\begin{aligned}
      \S_{trt}(t_e, r, t_{\tilde{e}})  = \| \t_e + \r^\circ - \t_{\tilde{e}} \|_{\ell2}^2 \ \ ,
\end{aligned}
\end{equation}
where $\t_e, \r^\circ, \t_{\tilde{e}} \in \Rbb^\ell$. The model returns a lower score if the two entity types is close under this relationship and a higher one otherwise. 

Fig.$\ref{fig:E2T-TRT}$ shows an illustration of E2T and TRT.

\subsection{Implementation for Entity Type Prediction}
Our framework can be used for entity type prediction in the following way. First, for each entity $e$ that appears in the testing set, a prediction by E2T is performed with:
\begin{equation}
    \label{E2T}
    \hat{t}_e = \mathop{\arg\min}_{t \in \Tcal} \ \ \S_{e2t}(e, t).
\end{equation}
In addition, a composite score (E2T+TRT) by connecting entity type instances and entity type triples with embedding model, which we call ConnectE \footnote{We also call it ConnectE-(E2T+TRT), and use ConnectE-(E2T+0) to denote E2T for uniformity in the experiments.}, is defined as follows:
\begin{equation*}
\begin{array}{l}
\begin{aligned}
    \S_{e2t+trt}(e,\ & t_e)= \lambda \cdot \S_{e2t}(e, t_e)  +  \\ 
    & (1- \lambda) \cdot \Big\{ \frac{1}{|P|} \sum_{t_{\tilde{e}} \in P}\S_{trt}(t_e, r, t_{\tilde{e}}) \\
    &\qquad \quad \  + \frac{1}{|Q|}\sum_{t_{\bar{e}} \in Q}\S_{trt}(t_{\bar{e}}, r, t_e) \ \Big\}  \ ,
\end{aligned}
\end{array}
\end{equation*}
where $\lambda$ is a hyperparameter for the trade-off. $P=\{t_{\tilde{e}} | t_{\tilde{e}} \in \Tcal ,  (e, r, \tilde{e}) \in \Dcal \}$ (i.e. given $e$ is head entity, $P$ is the set of all corresponding tail entities' types.), and $Q =\{t_{\bar{e}} | t_{\bar{e}} \in \Tcal ,  (\bar{e}, r, e) \in \Dcal\}$ (i.e. given $e$ is tail entity, $Q$ is the set of all corresponding head entities' types.). $|P|$ and $|Q|$ represent the total number of entity types in $P$ and $Q$ respectively. A prediction is performed with:
\begin{equation}
    \label{E2T+TRT}
    \hat{t}_e = \mathop{\arg\min}_{t_e \in \Tcal} \ \ \S_{e2t+trt}(e, t_e).
\end{equation}
Hence, our final composite model ConnectE-(E2T+TRT) favors predictions that agree with both entity type instances and global triple information in KGs.

\subsection{Optimization}
\label{learnig}
We use ranking loss algorithm for training ConnectE-(E2T+TRT), in which the parameter set $\Theta = \{\E, \T, \R^\star, \R^\circ, \M \}$. $\E, \T$ stand for the collection of all entities' and types' embeddings respectively. $(\R^\star, \R^\circ)$ denotes the collections of relationships' differentiated embeddings. The ranking objectives are designed to assign lower scores to true facts (including $(e,r,\tilde{e})$ triple facts, $(e,t)$ entity type instances and $(t_e,r,t_{\tilde{e}})$ type triples) versus any corrupt ones. 
We build three sub-objective functions, i.e., $\J_1, \J_2, \J_3$, and implement dynamic optimization strategy, i.e., fix a partial of parameters and only update the rest when minimizing each function.
(1) $\J_1$: We choose TransE (see \cites{Bordes:Translating}) to model triple facts as $\S(e,r,\tilde{e})$, in which we update the embeddings of entities ($\forall \e \in \E$) and the embeddings of relationships ($\forall \r^\star \in \R^\star$). 
(2) $\J_2$: We only update the embeddings of entity types ($\forall \t \in \T$) and projecting matrix $\M$, not the entities' embeddings that have been trained in $\J_1$. 
(3) $\J_3$: We only update the embeddings of relationships ($\forall \r^\circ \in \R^\circ$) while keeping the entity types' embeddings fixed. 
The training is performed using Adagrad \cite{Kingma2014Adam}. 
All embeddings in $\Theta$ are initialized with uniform distribution. The procedure, from $\J_1$, $\J_2$ to $\J_3$, is iterated for a given number of iterations. We have:

\begin{equation*}\footnotesize
\begin{aligned}
&\J_1 = \sum\limits_{\Dcal}{}  \sum\limits_{ \Dcal'}{} [\gamma_1 + \S  (e,r,\tilde{e}) - \S  (e',r,\tilde{e}') ]_+\ , \\
&\J_2 = \sum\limits_{ \Hcal}{}  \sum\limits_{ \Hcal'}{} [\gamma_2 + \S_{e2t}  (e,t_e) - \S_{e2t}(e', t_e') ]_+\ ,\\
&\J_3 = \sum\limits_{\Zcal}{}  \sum\limits_{\Zcal'}{} [\gamma_3 +  \S_{trt}(t_e, r, t_{\tilde{e}}) -  \S_{trt}(t_e', r, t_{\tilde{e}}') ]_+
\end{aligned}
\end{equation*}
$\gamma_1, \gamma_2, \gamma_3 >0$ are margin hyperparameters, and the corrupted datasets are built as follows:
\begin{equation*}\footnotesize
\begin{aligned}
\Dcal ' := &\{ (e',r,\tilde{e})|(e,r,\tilde{e}) \in \Dcal ,  e' \in \Ecal , e' \neq e \} \\ \cup &\{  (e,r,\tilde{e}')|(e,r,\tilde{e}) \in \Dcal , \tilde{e}' \in \Ecal , \tilde{e}' \neq \tilde{e}\} \ , \\ 
\Hcal ' := & \{ (e',t_e)|(e,t_e) \in \Hcal ,  e' \in \Ecal , e' \neq e \} \\ \cup &\{   (e,t_e')|(e,t_e) \in \Hcal ,  t_e' \in \Tcal , t_e' \neq t_e\} \ ,\\ 
\Zcal ' := & \{ (t_e',r,t_{\tilde{e}})|(t_e,r,t_{\tilde{e}}) \in \Zcal ,  t_e' \in \Tcal , t_e' \neq t_e\} \\ \cup &\{   (t_e,r,t_{\tilde{e}}')|(t_e,r,t_{\tilde{e}}) \in \Zcal ,  t_{\tilde{e}}' \in \Tcal , t_{\tilde{e}}' \neq t_{\tilde{e}}\}
\end{aligned}
\end{equation*}
$\Dcal,\Hcal$ are training datasets of triple facts and entity type instances in KG. $\Zcal$ is the training data of type triples, built by replacing entities in $\Dcal$ with their corresponding entity types. 

\section{Experiments}
\label{experiments}
\subsection{Datasets} 
We conduct the experiments on two real-world datasets ($\Dcal$) widely used in KG embedding literature, i.e. FB15k \cite{Bordes:Translating} and YAGO43k \cite{Moon:Learning}, which are subsets of Freebase \cite{Bollacker:Freebase} and YAGO \cite{Suchanek:Yago} respectively. They consist of triples, each of which is formed as (\emph{left entity, relationship, right entity}). We utilize two entity type data ($\Hcal$, each of it is formed as (\emph{entity, entity type})) built in \cite{Moon:Learning}, called FB15kET and YAGO43kET, in which the entity types are mapped to entities from FB15k and YAGO43k respectively. 

Moreover, we build new type triple datasets ($\Zcal$, each one in it is formed as (\emph{head type, relationship, tail type})), to train our model. They are built based on $\Dcal$ and $\Hcal$. First, for each triple $(e,r,\tilde{e}) \in \Dcal$, we replace the head and the tail with their types according to $\Hcal$. The generated datasets are called FB15kTRT(full) and YAGO43kTRT(full). Second, considering about the scalability of the proposed approach for full KGs, we further modify the generation method of type triples, which is the major training bottleneck. We discard newly generated ones with low-frequency (i.e. \#frequency = 1). After that the size of both FB15kTRT(full) and YAGO43kTRT(full) decreased by about 90\%, called FB15kTRT(disc.) and YAGO43kTRT(disc.) respectively. The statistics of the datasets are showed in Table $\ref{data-stcs}$. For saving space, we put more data processing details (include cleaning $\Hcal$, building $\Zcal$, etc.) on our \hytt{github} website.
\begin{table}[htb]\tiny
\caption{Statistics of $\Dcal, \Hcal, \Zcal$.}
\label{data-stcs}
\centering
\begin{tabular}[t]{c||c|c|c|c|c}
\toprule
{\bf Dataset }     & {\bf \#Ent} &{\bf \#Rel }&{\bf\#Train }& {\bf \#Valid} &{\bf \#Test }\\
\midrule
 {FB15k}  & 14,951 & 1,345 & 483,142 & 50,000 & 59,071 \\
 {YAGO43k}  & 42,335 & 37 &331,687 & 29,599 & 29,593 \\
\bottomrule
\end{tabular}
\begin{tabular}[t]{c||c|c|c|c|c}
\toprule
{\bf Dataset }     &{\bf \#Ent} &{\bf \#Type }&{\bf \#Train }&{\bf \#Valid }&{\bf \#Test }\\
\midrule
FB15kET &{ 14,951} & {3,851 }&{ 136,618 }&{ 15,749} &{ 15,780 }\\
\scriptsize{YAGO43kET}  &{ 41,723 }&{ 45,182 }&{ 375,853 }&{ 42,739 }&{ 42,750 }\\
\bottomrule
\end{tabular}
\begin{tabular}[t]{c||c|c|c|c|c}
\toprule
{\bf Dataset   }   &{\bf \#Type }& {\bf\#Rel} &{\bf \#Train} & {\bf Valid }&{\bf Test }\\
\midrule
FB15kTRT(full) &3,851& 1,345 & 2,015,338 &--&--\\
FB15kTRT(disc.) &2,060& 614 & 231,315 &--&--\\
YAGO43kTRT(full)  &45,128& 37 & 1,727,708 &--&--\\
YAGO43kTRT(disc.)  &17,910& 32 & 189,781 &--&--\\
\bottomrule
\end{tabular}
\end{table}

\subsection{Entity Type Prediction}
This task concentrates to complete a pair (\emph{entity, entity type}) when its type is missing, which aims to verify the capability of our model for inferring missing entity type instances.

\noindent{\textbf{Evaluation Protocol. }}
We focus on entity type prediction determined by Formula ($\ref{E2T}$) and (\ref{E2T+TRT}). We use ranking criteria for evaluation. Firstly for each test pair, we remove the type and replace it by each of the types in $\Tcal$ in turn. The function value of the negative pairs would be computed by the related models and then sorted by ascending order. We can obtain the exact rank of the correct type in the candidates. Finally, we use two metrics for comparison: (1) the mean reciprocal rank (MRR), and (2) the proportion of correct entities ranked in the top 1/3/10 (HITS@1/3/10)(\%). Since the evaluation setting of ``Raw'' is not as accurate as ``Filter'' \cite{Bordes:Translating}, we only report the experimental results with latter setting in this paper.
\begin{equation*}
\begin{aligned}
    MRR = \frac{1}{|C|}\sum_{i=1}^{|C|}\frac{1}{rank_i} \ , 
\end{aligned}
\end{equation*}
where $C$ is a set of test pairs, and $rank_i$ is the rank position of the true entity type for the $i$-th pair.

\begin{table*}[htb]\tiny
\centering
\caption{\textbf{Entity type prediction results.} Evaluation of different models on FB15kET and YAGO43kET.}
\label{table:etp}
\newcommand{\tabincell}[2]{\begin{tabular}{@{}#1@{}}#2\end{tabular}}
\begin{tabular}{l||c|c|c|c|c|c|c|c}
\toprule
{\bf DATASET }& \multicolumn{4}{c|}{{\bf FB15kET}} & \multicolumn{4}{|c}{{\bf YAGO43kET}}\\
\midrule
{\bf METRICS }&{\bf  MRR} &{\bf HITS@1}&{\bf HITS@3} & {\bf HITS@10} &{\bf MRR}  &{\bf HITS@1}&{\bf HITS@3} &{\bf HITS@10} \\
\midrule
RESCAL \cite{Nickel:A2011}  & 0.19&9.71&19.58&37.58 &0.08&4.24&8.31&15.31\\
RES.-ET   \cite{Moon:Learning}&0.24&12.17&27.92&50.72 &0.09&4.32&9.62&19.40\\
HOLE \cite{Nickel:Holographic}&0.22&13.29&23.35&38.16&0.16&9.02&17.28&29.25\\
HOLE-ET \cite{Moon:Learning} &0.42&29.40&48.04&66.73 &0.18&10.28&20.13&34.90\\
TransE \cite{Bordes:Translating} &0.45&31.51&51.45&73.93&0.21&12.63&23.24&38.93\\
TransE-ET \cite{Moon:Learning} &0.46&33.56&52.96&71.16&0.18&9.19&19.41&35.58\\
ETE \cite{Moon:Learning} &0.50&38.51&55.33&71.93&0.23&13.73&26.28&42.18\\
\midrule
\tabincell{l}{\bf ConnectE-(E2T+0)} &0.57 +- .00&45.54 +- .28 &62.31 +- .29&78.12 +- .12 &0.24 +- .01 & 13.54 +- .12 &26.20 +- .18 & 44.51 +- .09\\
\tabincell{l}{\bf ConnectE-(E2T+TRT)(disc.)}&{0.59 +- .01}&{48.54 +- .71}&{63.66 +- .39}&{78.27 +- .16}&{0.27 +- .01}& 15.1 +- .15 &29.14 +- .13&47.08 +- .09 \\
\tabincell{l}{{\bf ConnectE-(E2T+TRT)(full)}}&{\bf 0.59 +- .00}&{\bf 49.55 +- .62}&{\bf 64.32 +- .37}&{\bf 79.92 +- .14}&{\bf 0.28 +- .01}&{\bf  16.01 +- .12} &{\bf 30.85 +- .13}&{\bf 47.92 +- .07} \\
\bottomrule
\end{tabular}
\end{table*}

\noindent{\textbf{Implementation. }}
The results of entity type prediction are shown in Table $\ref{table:etp}$, where the results for the baselines are directly taken from original literature \cite{Moon:Learning}. We do not choose LM and PEM \cite{Neelakantan:Inferring} as baselines since they do not utilize triple knowledge, thus it is not fair to compare with them.
For training our model, we select the learning rate $\alpha$ $\in$ \{0.1, 0.05, 0.001\}, the margins $\gamma_1,\gamma_2,\gamma_3\in \{0.5, 1, 2, 5, 10\}$, the embedding dimension pairs $(\kappa, \ell)$ $\in$ \{(100, 50), (150, 75), (200, 100), (250, 125)\}, and the weight $\lambda$ $\in$ \{0.5, 0.65, 0.85, 0.95\}. We use negative sampling, and gradient descent with AdaGrad as our optimization approach to improve convergence performance. During the initialization process, each embedding vector of the entities, entity types and relationships is initialized with a random number following a uniform distribution $-\sqrt{6}/(m+n)$, where $n\in$ $\{$\#Ent, \#Type, \#Rel$\}$ and $m \in \{\kappa, \ell\}$. During the whole training process, we normalize the entity embeddings after each epoch.

We select the parameters based on MRR in valid dataset. The optimal configurations are: $\{\alpha=0.1, \gamma_1=\gamma_2=\gamma_3=2,\kappa =200, \ell=100, \lambda = 0.85 \}$ on FB15k/ET/TRT; $\{ \alpha=0.1, \gamma_1=\gamma_2=\gamma_3=1, \kappa=250, \ell=125, \lambda = 0.85 \}$ on YAGO43k/ET/TRT. We run 800 epochs on both datasets, and the batch size is 4096.

\noindent{\textbf{Experimental Results. }} We can see from Table $\ref{table:etp}$ that our ConnectEs outperform all baselines for entity type prediction in terms of all metrics on FB15kET and YAGO43kET. It confirms the capability of ConnectEs in modeling with local typing and global triple knowledge and inferring missing entity type instances in KGs. The model ConnectE-(E2T+TRT)(full) achieves the highest scores. 

\noindent {\bf Analysis. }(1) In E2T, we utilize a mapping matrix M which compresses entity embeddings into type embedding space, considering that entity type embedding represents common information of all the entities which belong to this type. The type embedding should be in a sharing subspace of entity embeddings. The experimental results of E2T compared with the baselines demonstrate that this assumption would be quite reasonable. 
(2) In E2T+TRT, we build new type-relation-type data, and then connect them with entity type instances. This approach provides more direct useful information to (weakly) supervise entity type prediction.  For example, given a fact that head entity \emph{Barack Obama} belongs to type \emph{/people/person} and the relationship \emph{born\_in}, we could make the best guess of the type of tail entity \emph{Honolulu} as \emph{/location/location}. Hence, the addition of type triples in ConnectE-{(E2T+TRT)} provides superior performance than ConnectE-{(E2T+0)}.
(3) Concerning about the scalability of our approach for big KGs, we utilize   FB15kTRT(disc.) and YAGO43kTRT(disc.) for prediction, the training time of which reduced by 90\% as the training data size decreased by 90\%. Moreover, the results of ConnectE-{(E2T+TRT)}(disc.) show that it's comparable with the best ConnectE-{(E2T+TRT)}(full).

 \begin{figure*}[htb]
\noindent\makebox[\textwidth][c] {
    \includegraphics[width=0.35\paperwidth]{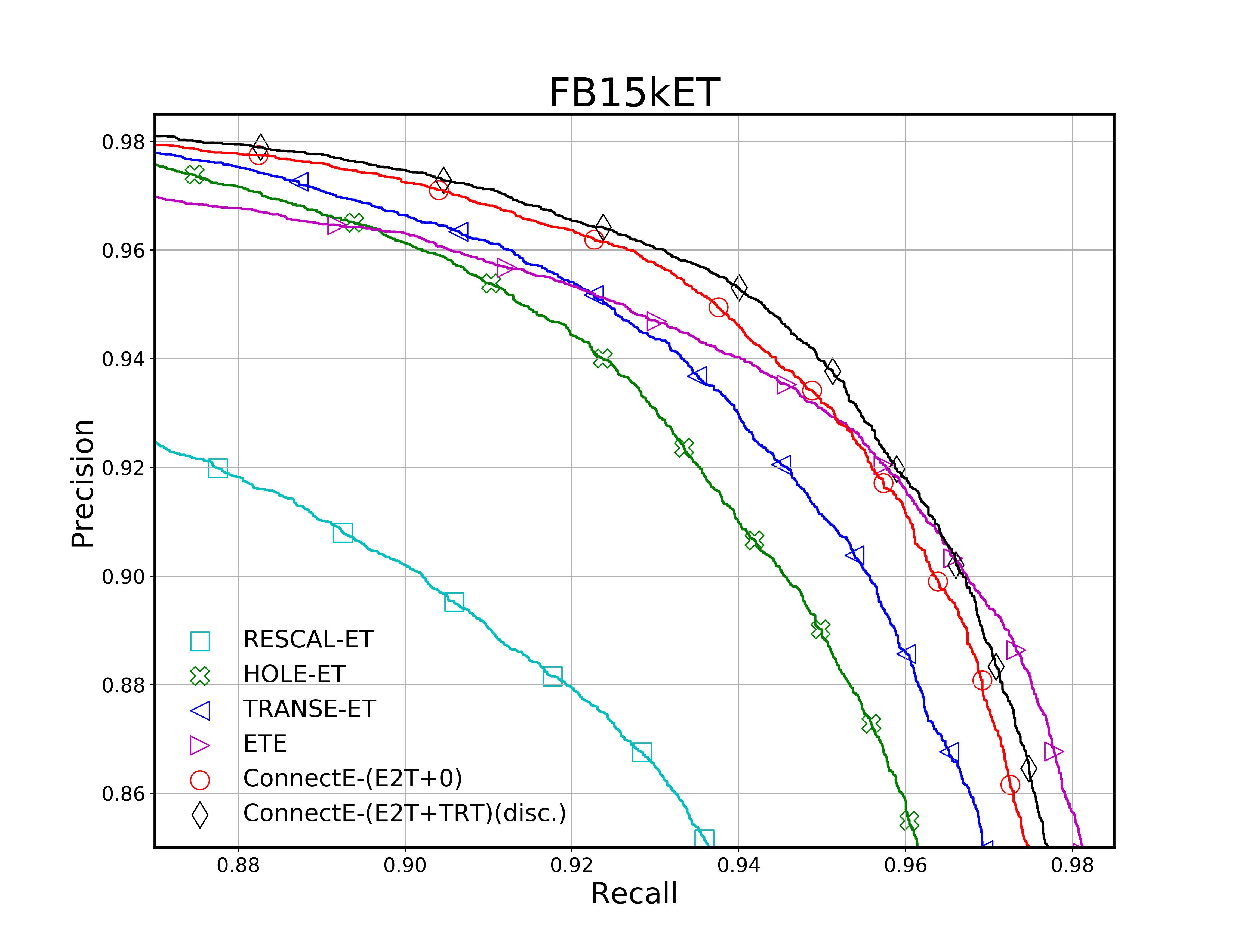}
    \includegraphics[width=0.35\paperwidth]{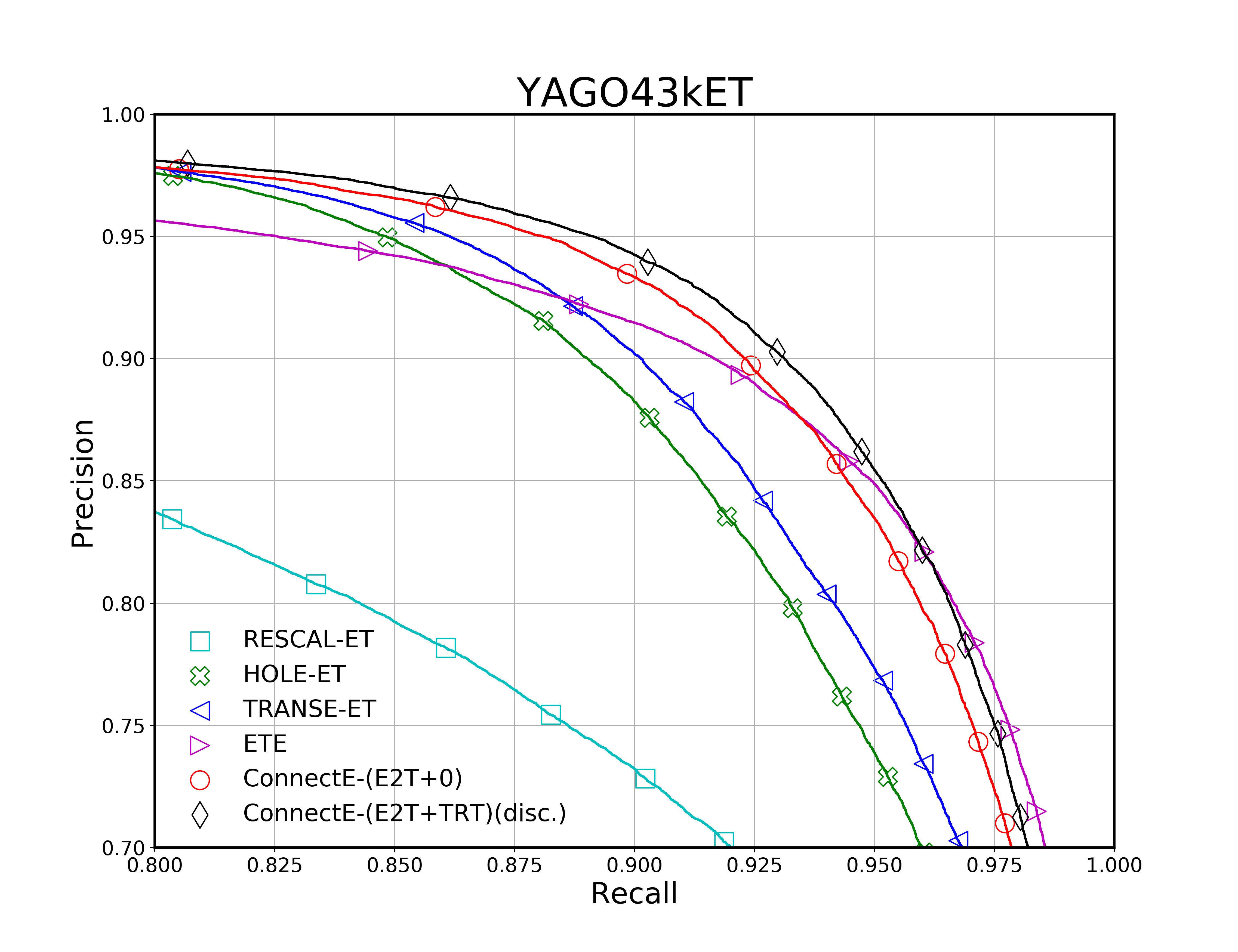} 
    }
\caption{\textbf{Entity type classification results (Precision/Recall Curve).} Evaluate on FB15kET, YAGO43kET.}
\label{fig:kgetnd}
\end{figure*}
\subsection{Entity Type Classification}
This task aims to judge whether each entity type instance in testing data holds or not, which could be viewed as a binary classification problem. 

\noindent{\textbf{Evaluation Protocol. }}
Since there are no explicit negative entity type instances in existing KGs, in order to create datasets for classification, we build negative facts by randomly switching type from entity type pairs in validation and testing set with equal number of positive and negative examples.
Inspired by the evaluation metric of triple classification in \cite{Socher:Reasoning}, we calculate the scores of all entity type instances based on model energy function, and rank all instances in testing set with these scores. Those instances with lower scores are considered to be true. We use {\bf precision/recall curves} to show the performances of all models.
Moreover, we also compare the {\bf accuracy} among different models. We first use validate set to find best threshold $\eta$. For instance, if the model score $\S_{e2t+trt}(e,t_e) \le \eta$ in classification, the entity type instance will be classified to be positive, otherwise to be negative. The final accuracy is based on how many facts are classified correctly.

\noindent{\textbf{Implementation. }}
We utilize the source codes and parameter settings of several baselines provided by \cite{Moon:Learning} for this task. The optimal parameter settings for our proposed models are: $\{ \alpha=0.1, \gamma_1=\gamma_2=\gamma_3=2,\kappa =200, \ell=100, \lambda = 0.85 \}$ on FB15kET; $\{ \alpha=0.1, \gamma_1=\gamma_2=\gamma_3=1, \kappa=250, \ell=125, \lambda = 0.85 \}$ on YAGO43kET. In both datasets, we learn all the training data for 800 epochs and the batch size is 4096.
After training, we firstly draw PR-curves with dynamic thresholds. We select the best threshold based on the accuracy in valid dataset, which is used to calculate the accuracy in test dataset.
 
\noindent{\textbf{Experimental Results. }}
We draw the PR-curves for type classification task on both datasets in Fig.\ref{fig:kgetnd}. 
Note that we only report the results of ConnectE-(E2T+TRT)(disc.) not ConnectE-(E2T+TRT)(full), since the learning speed of the former is much more faster than the latter and its results are close to the best results of the latter. We can see from Fig.\ref{fig:kgetnd} that when the recall rate is between 0.88 $\sim$ 0.97, ConnectE-(E2T+TRT)(disc.) model could achieve the highest precision rate on FB15kET. In other ranges, our ConnectE-(E2T+TRT)(disc.) model also shows comparable performance. The result is consistent on YAGO43kET. Specifically, ConnectE-(E2T+TRT)(disc.) achieves the best F1 score of 94.66\% when recall = 94.27\% and precision = 95.05\% on FB15kET. Also, ConnectE-(E2T+TRT)(disc.) surpasses other models and gets F1 score of 92.13\% when precision = 93.18\% and recall = 91.11\% on YAGO43kET. It confirms the capability of our model, for they could not only infer missing types in KGs, but also perform well in KG entity type classification.

Table $\ref{table:bc}$ demonstrates the evaluation accuracy results of entity type classification, from which we can observe that: 
(1) On FB15kET, ConnectE-(E2T+TRT)(disc.) achieves the best accuracy score (94.49\%).  Compared to the mostly related model ETE, our model shows 0.48\% absolute performance improvement. On YAGO43kET, ConnectE-(E2T+TRT)(disc.) model outperforms other models as well. The improvement of our model compared to ETE is almost 1.51\%. 
(2) Comparing to the improvement on YAGO43kET, the advantage ConnectE-(E2T+TRT)(disc.) has over ConnectE-(E2T+0) in this task on FB15kET seems to be insignificant, which indicates that the type triples in FB15kTRT have fewer contribution on entity type classification than ones in YAGO43kTRT. It may be partially caused by the fact that the number of relations in YAGO43k (\#Rel=37) is far less than that in FB15k (\#Rel=1,345), which could considerably influence the effectiveness of the type-relation-type training set. Due to the rareness of relationships in YAGO43k, each entity usually connects with a large number of other entities through one single relationships, which means that the magnitude of $|P|$ and $|Q|$ in the composite model scoring function are large. After averaging in ConnectE-(E2T+TRT)(disc.), it could achieve more stable and significant results on YAGO43kET. 
\begin{table}[htb]\tiny
\caption{\textbf{Entity type classification results (accuracy).} 
}
\label{table:bc}
\newcommand{\tabincell}[2]{\begin{tabular}{@{}#1@{}}#2\end{tabular}}
\begin{center}
\begin{tabular}[t]{l||c|c}
\toprule
{\bf Dataset}      & \tabincell{c}{{\bf FB15kET} } &\tabincell{c}{{\bf YAGO43kET }}\\
\midrule
RESCAL-ET & 90.02\% & 82.28\%     \\
\midrule
HOLE-ET & 93.23\% & 90.14\%     \\
\midrule
TransE-ET & 93.88\% &  90.76\%    \\
\midrule
ETE & 94.01\% &  90.82\%    \\ 
\midrule
{\bf ConnectE (E2T+0)} &  94.45\% &  91.78\%  \\
{\bf ConnectE (E2T+TRT)(disc.)} &{\bf 94.49}\% & {\bf  92.33}\% \\
\bottomrule
\end{tabular}
\end{center}
\end{table}
\subsection{Case Study}
Table $\ref{table:cs}$ shows the examples of entity type prediction by our model from \emph{FB15k/ET/TRT}, which demonstrate our motivation of Mech. $\ref{machanism-TRT}$ that head type and tail type really maintain the relationship between head entity and tail entity. Given entity \emph{Peter Berg}, TRT can find HITS@1 type prediction \emph{/people/person} for it via the existing entity type assertion (\emph{New Youk, /location/location}) and the relationship (\emph{/loc./loc./people\_born\_here}) between them, i.e. $\vec{{\emph{Peter Berg}}} - \vec{{\emph{New York}}} + \vec{{\emph{/location/location}}}$= $\vec{{\emph{/people/person}}}$.

\begin{table}[htb]\tiny
\centering
\caption{\textbf{Entity type prediction examples.} Extraction from \emph{FB15k/ET/TRT}.}
\label{table:cs}
\newcommand{\tabincell}[2]{\begin{tabular}{@{}#1@{}}#2\end{tabular}}
\begin{tabular}{l|c|c|c|c|c}
\toprule
&\multicolumn{2}{c|}{\tabincell{c}{ \textbf{\texttt{Type prediction:}} \\ \textbf{\texttt{HIT@1}}} } & \textbf{\texttt{Rel}} &\multicolumn{2}{c}{ \tabincell{c}{\textbf{\texttt{Tail type}}}} \\
\midrule
\multirow{3}*{1} &\multicolumn{2}{c|}{\tabincell{c}{\textbf{\texttt{type=?}}\\ \textbf{\emph{/people/person}}}} & \multirow{3}*{\tabincell{c}{\emph{/location/location/}\\ \emph{people\_born\_here}}} &\multicolumn{2}{c}{ \textbf{\emph{ /location/location}} } \\
\cline{2-3} \cline{5-6}
&\multirow{2}*{\tabincell{c}{head\\ entity}}&\emph{Peter Berg}& & \emph{New York} & \multirow{2}*{\tabincell{c}{tail\\entity}}\\
\cline{3-3} \cline{5-5}
& &\emph{Gus Van Sant}& &\emph{Louisville} & \\
\midrule
\multirow{3}*{2} &\multicolumn{2}{c|}{\tabincell{c}{ \textbf{\texttt{type=?}}\\ \emph{\bf /americancomedy/movie}  }} & \multirow{3}*{\tabincell{c}{\emph{/film/film/} \\ \emph{directed\_by}}} &\multicolumn{2}{c}{ \textbf{\emph{ /film/director}} } \\
\cline{2-3} \cline{5-6}
&\multirow{2}*{\tabincell{c}{head\\ entity}}&\emph{Very Bad Things}& & \emph{Peter Berg} & \multirow{2}*{\tabincell{c}{tail\\entity}}\\
\cline{3-3} \cline{5-5}
& &\emph{Rush Hour}& &\emph{Brett Ratner} & \\
\midrule
\multirow{3}*{3} &\multicolumn{2}{c|}{\tabincell{c}{\textbf{\texttt{type=?}} \\ \textbf{\emph{/medicine/disease}  }}} & \multirow{3}*{\tabincell{c}{\emph{people/cause\_of}\\ \emph{\_death/people}}} &\multicolumn{2}{c}{ \textbf{\emph{ /people/person}} } \\
\cline{2-3} \cline{5-6}
&\multirow{2}*{\tabincell{c}{head\\ entity}}&\tabincell{c}{\emph{Myocardial}\\ \emph{infarction}}& & \emph{Dick Clark} & \multirow{2}*{\tabincell{c}{tail\\entity}}\\
\cline{3-3} \cline{5-5}
& &\tabincell{c}{\emph{Pancreatic} \\ \emph{cancer}}& &\emph{John Hurt} & \\
\bottomrule
\end{tabular}
\end{table}
\section{Conclusion and Future Work}
\label{conclusion}
In this paper, we described a framework for leveraging global triple knowledge to improve KG entity typing by training not only on (\emph{entity, entity type}) assertions but also using newly generated (\emph{head type, relationship, tail type}) type triples. Specifically, we propose two novel embedding-based models to encode entity type instances and entity type triples respectively. 
The connection of both models is utilized to infer missing entity type instances. The empirical experiments demonstrate the effectiveness of our proposed model.
Our modeling method is general and should apply to other type-oriented tasks. Next, we are considering to use this framework to conduct KG entity type noise detection. 

\section*{Acknowledgments}
The authors would like to thank all anonymous reviewers for their insightful comments. We also want to thank Zhiyuan Liu (Tsinghua University) and Linmei Hu (BUPT) for their useful suggestions and comments on early drafts. This work was supported by the National Natural Science Foundation of China under Grant No.61922085, 61906159, the Sichuan Science and Technology Program under Grant No.2018JY0607, the Fundamental Research Funds for the Central Universities under Grant No.JBK2003008, Fintech Innovation Center, and Financial Intelligence and Financial Engineering Key Laboratory of Sichuan Province.

\bibliography{acl2020}
\bibliographystyle{acl_natbib}

\end{document}